# Ant Colony Algorithm for the Weighted Item Layout Optimization Problem

Yi-Chun Xu[1], Fang-Min Dong[1], Yong Liu[1], Ren-Bin Xiao[2], Martyn Amos[3,*]

[1]Institute of Intelligent Vision and Image Information, China Three Gorges University, China.

[2]Institute of Systems Engineering, Huazhong University of Science and Technology, China.

[3]Department of Computing and Mathematics, Manchester Metropolitan University, United Kingdom.

[*]Corresponding author. Email: m.amos@mmu.ac.uk.

*Abstract*—**This paper discusses the problem of placing weighted items in a circular container in two-dimensional space. This problem is of great practical significance in various mechanical engineering domains, such as the design of communication satellites. Two constructive heuristics are proposed, one for packing circular items and the other for packing rectangular items. These work by first optimizing object placement order, and then optimizing object positioning. Based on these heuristics, an ant colony optimization (ACO) algorithm is described to search first for optimal positioning** *order***, and then for the optimal** *layout***. We describe the results of numerical experiments, in which we test two versions of our ACO algorithm alongside local search methods previously described in the literature. Our results show that the constructive heuristic-based ACO performs better than existing methods on larger problem instances.**

I. INTRODUCTION

*Cutting and Packing* (C&P) problems are optimization problems with practical significance to manufacturing industries (e.g. when cutting glass, wood, and leather) or transportation (Dyckhoff, 1990). *Layout optimization problems*, such as the placement of cells in VLSI design (Tang and Yao, 2007) or the layout of articles in a newspaper (Gonzalez *et al.*, 2002), are related to C&P problems. In this domain, small items must be placed in a "container" without overlap, according to some certain objectives. In general, these optimization problems are NP-hard (Dyckhoff, 1990). When constraints are attached to them, they usually become even more complex.

Existing approaches for solving C&P problems include *local search* heuristics and *constructive* heuristics. Local search heuristics (e.g., genetic algorithms (Tang and Yao, 2007), simulated annealing (Gonzalez *et. al*, 2002; Cagan *et al.*, 1998), and gradient descent method (Wang *et al.*, 2002)) search the neighborhood of a given starting solution and improve it continuously, until a near-optimal solution is found. Constructive heuristics place the items one-by-one in a certain sequence (order); when the final item is placed, a good solution is thus constructed (Dowsland *et. al*, 2002; Wu *et al*, 2002). Several papers have successfully combined these two approaches; they first use a constructive heuristic to place items in a given sequence, and then apply a local search to improve the layout (Harwig, 2003; Sokea & Bing, 2006).

In this paper, we study the problem of packing weighted items in a circular container; the so-called W*eighted Item Layout* problem (WIL). The WIL was first proposed in (Teng *et al.*, 1994); as described, items may not overlap, and the whole system must be kept *balanced* in terms of its distribution of mass. One significant real-world application of WIL may be found in the design of communication satellites (Teng *et al*, 2001; Zhang *et al.,* 2008), where balancing the distribution of mass is critical to the stability of a large rotating cylinder. In order to reduce complexity, the problems we consider here are limited to two-dimensional space. Two variants of the problem are considered here; the first places weighted *circles* (Fig1-a), and the second places weighted *rectangles* (Fig1-b). We name these variants the *Weighted-Circle Layout* problem (WCL) and *Weighted-Rectangle Layout* problem (WRL) respectively.

Several approaches to the WIL have already been proposed, and most of these are local search heuristics (Teng *et al.,* 1994; Teng *et al*., 2001; Zhang *et al.*, 2008; Xiao *et al.*,2007; Huang & Chen, 2006; Xu *et al*., 2007a). The main strategy of these local search heuristics is to first randomly place objects in the plane, and then gradually move them to new positions in order to decrease overlap and imbalance of mass within the system. When the items are circular, the overlap between them can be modeled as potential energy, and thus items may be moved according to the virtual "force" required to decrease energy (Wang *et al.*, 2002). Local search heuristics show reasonable performance for the WCL (Xiao, *et al.*, 2007; Huang & Chen, 2006), but when the same heuristics are applied to the WRL (or more complex problems in three-dimensional space), their performance quickly suffers (Xu *et al*., 2007). The main reason for this degradation is that the notion of a "good" layout is related not only to the *positions* of rectangles, but also to their *orientations* (which are not considered by local methods). We therefore focus on *constructive* heuristics for both the WCL and WRL.

For C&P problems, the most important components of a constructive heuristic are the *positioning rules*, i.e., how and where to place each item. By the positioning rules developed in (Xu *et al.*, 2007; Xu & Xiao, 2008; Xu *et al.,* 2010), we may place the weighted items one-by-one in a certain order, and then a good layout is generated. This method was named the *Order-based*

*Positioning Technique* (OPT). However, different orders will yield layouts of differing quality quality; the key problem is how to find the best *order* to improve the quality of a layout.

In the present paper, we develop ant colony optimization (ACO) algorithms to search for the optimal positioning order. ACO is a *meta-heuristic* for combinatorial optimization problems, and has gained increasing use in the last decade. The first version of the ACO algorithm, named AS (ant system), was proposed by Dorigo as a solution to the Traveling Salesman Problem (TSP) (Dorigo, 1992; Dorigo *et al*., 1996; Blum & Dorigo, 2004; Dorigo & Stützle, 2004). The authors based their algorithm on the observation that real ant colonies can quickly find a shortest path from their nest to a given food source, using chemical signalling based on *pheromones*. In the AS, Dorigo designed a colony of artificial ants, where each ant can build solutions by contributing to an artificial pheromone trail. The trail can also be externally reinforced according to the quality of the solution it represents. This mechanism forms a positive feedback loop, driving the system towards optimality. Since the invention of the AS, many improvements have been proposed to the original algorithm, and ACO has now been applied to many optimization problems (Dorigo & Stützle, 2004; Stützle & Hoos, 1997; Juang & Hsu, 2009) .

Some versions of ACO have already been applied to C&P problems. The first such application was to the one-dimensional bin-packing problem (1-BPP) (Levin & Ducatelle, 2004). The authors used ACO to learn which items are likely to be grouped together by their length. They reported that their ACO method outperformed other meta-heuristics such as genetic algorithms. )In (Brugger *et al*. 2004) the authors also reported a version of ACO for the 1-BPP, but they related the length of item to the space left in the bin. On the benchmarks, their ACO algorithm gave better results than (Levin & Ducatelle, 2004). Two-dimensional packing problems have also been solved by ACO; (Thiruvady *et al*., 2008) reports an ACO algorithm for the strip packing problem. Their method was based on a constructive heuristic ("Bottom-Left" method), and their method was used to learn the packing order. In (Burke & Kendall, 1999) the authors apply ant systems to the nesting problem, when the items are irregular. They used the "No-Fit-Polygon" to compose the constructive heuristic, and the ACO was also applied to learn the packing order.

Based on a survey of previous work, it seems that when applying ACO on C&P problems, it is necessary to first develop a constructive heuristic. In this paper, after introduction of the construction heuristic OPT, we develop two version of the ACO and compare their performance on both the WCL and the WRL. The rest of the paper is organized as follows: in Section II, we first define the mathematical foundations of the problem. We then introduce, in Section III, the constructive heuristic OPT for the WCL and WRL. The ACO algorithm is presented in Section IV, and we report the results of numerical experiments in Section V. We conclude in Section VI with a discussion and suggestions for future work.

## II. MODELS FOR WEIGHTED CIRCLE LAYOUT AND WEIGHTED RECTANGLE LAYOUT

Both problems are concerned with the placement of objects within some containing circle, the idea being to minimise the container's radius (whilst minimising mass imbalance, which we consider later). For both problems, we use $R$ to denote the radius of the containing circle. Suppose there are $n$ items to be placed and their masses are $m_1, m_2, \ldots, m_n$. For the WCL, the sizes of the circular items are denoted by their radii, $r_1, r_2, \ldots, r_n$. For the WRL, the size of each rectangular item is denoted by the lengths of its two adjacent edges, such as $(a_1, b_1), (a_2, b_2), \ldots, (a_n, b_n)$. We also use $r_1, r_2, \ldots, r_n$ to denote the radii of the enveloping circle of each rectangular item, such that $r_i = \frac{1}{2}\sqrt{a_i^2 + b_i^2}$.

In two-dimensional space, we use $(x_i, y_i)$ to denote the position of item $i$. For a circular item, $(x_i, y_i)$ is sufficient to define the position of item $i$, but for a rectangular item, we require additional information; the orientation $z_i$ is used together with $(x_i, y_i)$.

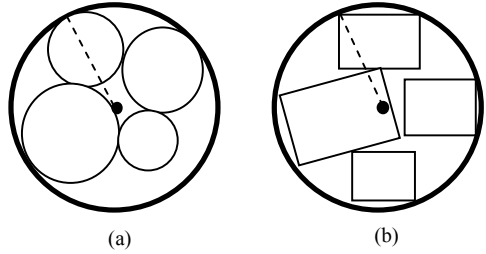

(a)      (b)

Figure 1. Illustration of WCL and WRL. (a) WCL (b) WRL.

By summarizing the models provided in the literature, we describe the three main objectives for each problem in the following subsections.

### A. Requirements of Weighted Circle Layout

#### 1. Non-overlapping items

This is the fundamental requirement, in that no two circles may overlap. Checking the condition for zero overlap between two circles is straightforward, in that we only need to check whether (1) is satisfied for each pair $i, j$.

$$\sqrt{(x_i - x_j)^2 + (y_i - y_j)^2} \geq r_i + r_j \qquad (1)$$

#### 2. Compactness of the layout

We require that items are placed in a "compact" layout. The compactness of the layout is measured by the radius of the enveloping circle, as illustrated in Fig1-a and Fig1-b. One objective of our packing is to obtain the smallest possible enveloping circle. The radius of the enveloping circle can be calculated for WCL by (2).

$$R\_envelope = \max_{1 \leq i \leq n}(r_i + \sqrt{x_i^2 + y_i^2}) \qquad (2)$$

*3. Balance of the system*

Another objective is to minimize the imbalance generated by weighted items, defined by (3):

$$imbalance = \sqrt{(\sum_{i=1}^{n} m_i x_i)^2 + (\sum_{i=1}^{n} m_i y_i)^2} \qquad (3)$$

In recent work (Xiao *et al.*, 2007; Xu *et al.*, 2007a), we showed that the optimization process can actually *benefit* from the zero-imbalance requirement. Under this stricter constraint, the mass center of the system is arranged at the center of the containing circle. Then the radius of the enveloping circle (to thus measure the compactness) is replaced by (4):

$$R\_envelope = \max_{1 \leq i \leq n}(r_i + \sqrt{(x_i - \frac{\sum_{i=1}^{n} m_i x_i}{\sum_{i=1}^{n} m_i})^2 + (y_i - \frac{\sum_{i=1}^{n} m_i y_i}{\sum_{i=1}^{n} m_i})^2}) \qquad (4)$$

It should be noted that we only consider *static* imbalance. *Dynamic* balance is not considered, as in (Teng *et al.*, 1994), because in the 2-dimensional case, when the static imbalance becomes zero, the dynamic imbalance will also get to zero.

*B. Requirements of Weighted Rectangle Layout*

*1. Non-overlapping items*

Although it is more difficult to judge the overlap of two rectangles compared with circles, it is still a straightforward issue in computational geometry. The sufficient and necessary conditions for no overlap between two rectangles *i* and *j* are:

1. Any edge of *i* does not intersect with any edge of *j*, and

2. Any vertex of *i* is not contained in *j*, and vice versa.

Condition 1 should be tested for every edge pair (*i*, *j*). After condition 1 is satisfied, only one vertex of rectangle *i* and one vertex of rectangle *j* should be checked to consider condition 2.

In general, the orientation of a rectangle is not a fixed value. But when we limit the orientation to be either 0 or 90 degrees, the conditions for no overlap become easier to check. Using (*xmin*, *ymin*), (*xmax*, *ymax*) to denote the left-bottom vertex and top-right vertex, we only need to check whether the following is satisfied (as illustrated by Fig 2):

$xmin_i > xmax_j$ or $xmax_i < xmin_j$ or $ymin_i > ymax_j$ or $ymax_i < ymin_j$

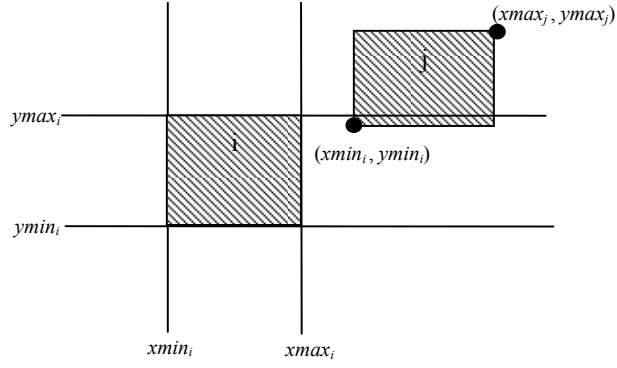

Figure 2. Relative position of two rectangles with fixed orientation.

*2. Compactness of the layout*

The enveloping circle illustrated in Fig1-b is again used to measure compactness. The calculation of the radius of the enveloping circle for WRL is a little more complex than in WCL, but it is still relatively easy. We calculate the distances from each vertex of the rectangles to the weight center of the container, and select the longest distance as the radius of the enveloping circle. We denote the set of four vertices of rectangle $i$ to be $\{v_{i,j} \mid j=1,2,3,4\}$, $i=1,2,..n$, and $d(a, b)$ to be the Euclidean distance between point $a$ and $b$. Suppose the center of the container is $o$, then the radius of the enveloping circle is defined as (5):

$$R\_envelope = \max_{i=1,2,\ldots,n} \max_{j=1,2,3,4} (d(v_{i,j}, o)) \qquad (5)$$

*3. Balance of the system*

Since the definition of imbalance of WRL is the same as (3), and the imbalance is also required to be zero, then the center of the enveloping circle should also be placed at the mass center of the system. Accordingly, the radius of the enveloping circle can also be found with reference to the mass center, where we replace the $o$ in equation (5) with a term describing the mass center of the system.

*C. Optimization for Weighted Item Layout*

Based on what we have described, WIL becomes a single objective optimization problem:

- Under the constraint of no overlap between any items, the enveloping circle centered at the mass center of the system should be minimized.

III. ORDER-BASED POSITIONING TECHNIQUE

The Order-based Positioning Technique (OPT) OPT is a constructive method for WIL. According to the defined positioning rules, objects are placed one-by-one in a given order, and the compactness and balance of the layout are assessed during this positioning. In general, if the sizes and masses of the items are different, the positioning order will affect the quality of the final layout. Without loss of generality, we consider a positioning order of (1, 2,…, $n$).

The main idea of OPT is that when we add a new item to a *partial layout* (the items already placed), we should find a position which will lead to a new partial layout with a minimally-sized enveloping circle. This is a *greedy* policy. The greedy policy can not guarantee a globally optimal layout, but can always provide a reasonable layout good. During OPT, we address two questions:

1. How and where to place the first item(s);

2. How and where to place the $i$-th item after the previous $i$-1 items.

We now describe OPT for WCL and WRL in two separate subsections.

*A. OPT for Weighted Circle Layout*

The rules to decide the position of a new circle in a partial layout are related to the compactness defined by (4). It is obvious that the tangent of two circles makes for the most compact layout if we have only two items, so it is reasonable to insist that a circle should be tangential to other circles when it is placed. Since there exist too many positions for a circle to be tangential to only one other circle (one circle can "run around" the other circle), we require that the new circle be tangential to *two* existing circles, such that there are only two available positions. The packing rules are described as follows:

1. Place circle 1 and circle 2 at position (-$r_1$, 0), ($r_2$, 0). From this rule, Circle 1 is placed tangential to circle 2.

2. From circle $i$>=3, we require circle $i$ to be tangential to two existing circles $p$ and $q$, where $p<i$ and $q<i$.

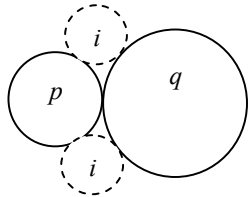

Figure 3. Positioning of a circle.

There are two available positions for circle *i* to be tangential to *p* and *q*. (Fig 3). Although these two positions can guarantee no overlap between *i* and *p* or *q*, overlap may still exist between *i* and other existing circles, so the validity of the two positions is checked. Given that there are $(i-1)(i-2)/2$ pairs of *p* and *q* altogether, then we may have, at most, $(i-1)(i-2)$ positions. From these positions, the one yielding the minimal R_envelope is chosen as the position of *i*.

If the calculation (with validation check) for one position is assumed to take unit time, and we assume $n>=3$, the time complexity of the OPT for WCL may be calculated as (6):

$$T(n) \sim \sum_{i=1}^{n}(i-1)(i-2) = O(n^3) \qquad (6)$$

B. *OPT for Weighted Rectangle Layout*

In the WCL, the *positions* of the all circles define the layout. In WRL, however, the *orientation* of each rectangle must also be provided along with its position. The less vacant space there is in a layout, the "better" the layout. So, when placing a rectangle, we try to find a position *and* orientation which minimise vacant space. The packing rules for WRL are described as follows:

1. The position and orientation of the first rectangle is directly set to $(x_1, y_1, z_1)=(0, 0, 0)$.

2. When positioning the rectangle *i*, we consult one existing rectangle *j* as illustrated in Fig 4. At first, we require one edge of rectangle *i* to be in contact with one edge of rectangle *j*. By this requirement and rule 1, we can infer that the orientation for any rectangle should be 0 or 90 degrees. Second, along the touching edges, we only consider two positions, as illustrated in Fig4-a, where the left or right vertices of the two touching edges coincide.

We believe these packing rules generate good layouts, since fewer *stiles* are generated. A stile introduces vacant space into a layout. In Fig 4-a, the partial layout has only one stile, but in Fig 4-b it has two.

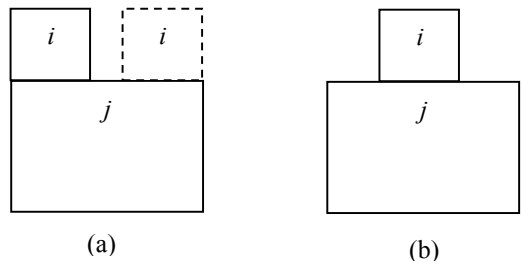

(a)        (b)

Figure 4. Illustration of the positioning rules for WRL (a) Item *i* is placed as defined, and generates one stile, (b) Item *i* generates two stiles.

As illustrated in Fig 5, rectangle *i* has 16 available positions around rectangle *j*. We also check the validity of the 16 positions, since *i* may overlap with other rectangles. Because there are *i*-1 existing rectangles, rectangle *i* has 16(*i*-1) available positions at most. Since rectangle *i* needs only one position, we also adopt a greedy strategy to choose the position which yields a minimal enveloping circle.

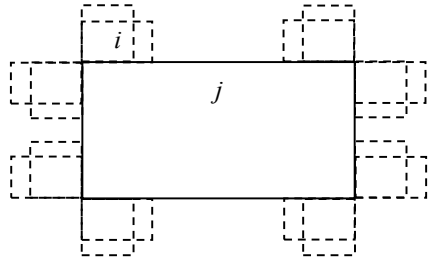

Figure 5. 16 positions for a rectangle.

Taking one position calculation as a unit time expense, we use (7) to estimate the time complexity of OPT.

$$T(n) \sim O(16 \times 1 + 16 \times 2 + ... + 16 \times (n-1)) = O(n^2) \qquad (7)$$

*C. Disadvantages of OPT*

OPT can generate a layout using a placing order. However, according to the constraints of the OPT, the layout can not always be optimal, especially when there are only a few items. This is easily demonstrated by Fig 6; layouts (c) and (d) are better than (a) and (b), but OPT is only capable of generating (a) and (b).

Fortunately, as the problem size increases, the advantages of OPT become apparent. In practice, we often need a satisfactory *near*-optimal solution, if we cannot find the optimal ones.

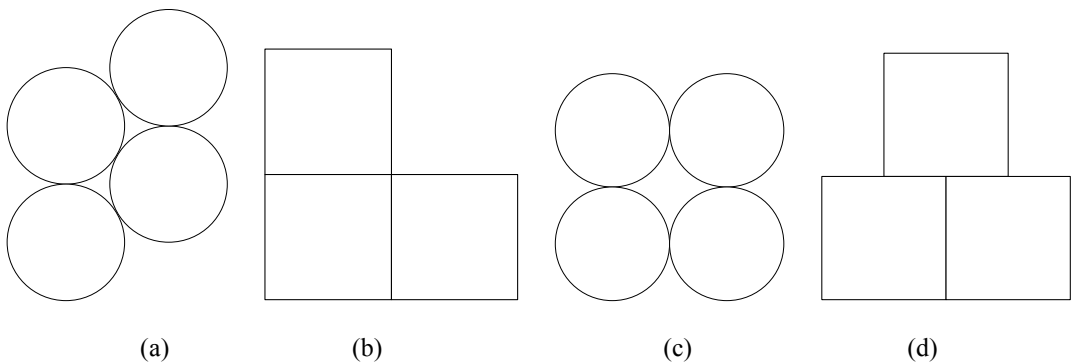

(a)  (b)  (c)  (d)

Figure 6. Illustration of OPT not able to provide the optimal layouts.(a) and (b) are the layout by OPT for WCL and WRL, where there are 4 items with same sizes and weights. (c) and (d) are two better layout than (a) and (b).

## IV. ANT COLONY OPTIMIZATION

OPT generates a layout according to a given positioning order. But for $n$ items with different sizes and different masses, there exist $n!$ positioning orders. Since different positioning orders can lead to different layouts, we should search for a positioning *order* which yields an optimal layout. This kind of problem is sometimes called a *permutation constrained problem*, where an optimal permutation of $(1, 2, …, n)$ is required. One famous permutation constrained problem is the TSP (Travelling Salesman Problem), whereby a salesman should visit $n$ cities only once, along the shortest path. Such problems quickly become intractable for even small values of $n$. ACO is a type of meta-heuristic well-suited to permutation constrained problems. In this Section, we provide an ACO algorithm to search for the optimal positioning order for WIL.

### A. Standard ACO for TSP

ACO is inspired by the behavior of real ants. As a kind of social insect, a single ant has very limited ability, but, through cooperation, the colony can perform very complex tasks. Ethnologists observe that, although the ant is almost blind, the colony as a whole can find a shortest path from their nest to the food source. When one ant moves on a path, it deposits *pheromone* (a chemical signalling molecule) on the ground. The other ants follow the (perhaps many) paths defined by the pheromone, where the path with the strongest concentration of pheromone is more likely to be chosen. Because the ants on the shorter path return more quickly, then the pheromone on the shorter path is *reinforced* and as a result it will attract more ants again. This is a positive feedback loop, which quickly leads to the emergence of a shortest path.

Dorigo formalised the above mechanics and invented the ant colony optimization method for TSP (Dorigo *et al*., 1996). Given $n$ cities $(1, 2, … n)$ with the distances $d_{ij}$ between any two city $i$ and $j$, the salesman should find the shortest closed tour, such that each city is visited once.

Suppose $L$ ants walk on the graph of $n$ cities. In ACO, each ant $k$ in city $i$ will move to city $j$ with probability $p_k(i,j)$,

$$p_k(i,j) = \begin{cases} \dfrac{\tau(i,j)^\alpha \cdot \eta(i,j)^\beta}{\sum_{s \in Allowed_k} \tau(i,s)^\alpha \cdot \eta(i,s)^\beta}, & \text{if } j \in Allowed_k \\ 0, & \text{otherwise} \end{cases} \quad (8)$$

We define $\tau(i,j)$ as the pheromone level on the edge $(i,j)$; $\eta(i,j)$ is the heuristic information guiding the ant, which in TSP is often set to $1/d_{ij}$, since shorter edges are preferred. $Allowed_k$ is a set includes the cities not visited by ant $k$. The parameters $\alpha$ and $\beta$ are used to define the relative importance of the pheromone and the heuristic information, which may be adjusted by the user.

After each ant performs a complete tour, the pheromone trails are updated. The pheromone on all edges "decays" at some rate $\rho$ $(0<\rho<1)$, and the pheromone on the edges traversed by ants is then *reinforced*.

$$\tau(i,j) \leftarrow \rho \cdot \tau(i,j) + \sum_{k=1}^{m} \Delta\tau_k(i,j) \qquad (9)$$

Where

$$\Delta\tau_k(i,j) = \begin{cases} \dfrac{Q}{L_k}, & \text{if } (i,j) \text{ walked by ant k} \\ 0, & \text{otherwise} \end{cases} \qquad (10)$$

$Q$ is a const, and $L_k$ is the length of the tour by ant $k$.

After the update of the pheromone trails, the next iteration begins. The ACO algorithm is illustrated in Fig 7.

```
ACO algorithm for TSP
{
    Initialize the parameters and the pheromone trail.
    For (iteration = 1 to MAX)
    {
        For (ant k=1 to m)
            While (any city is not visited by k)
                Determine next city with probability p_k(i, j) as defined by (8)
        Update the pheromone on the walked paths by (9) and (10)
    }
    Output the best tour and other data
}
```

Figure 7. ACO pseudo-code.

*B. Application of ACO to optimization problems*

As a robust and versatile meta-heuristic, ACO may be applied to solve many other combinatorial optimization problems, such as the Quadratic Assignment Problem and Job-shop Scheduling Problem (Dorigo & Stützle, 2004). The main assumptions, when applying ACO to a problem are that:

1. Each path followed by an ant may be mapped to a candidate solution to the given problem;

2. After a candidate solution is generated, the amount of pheromone to deposit on the path is proportional to the *quality* of the solution;

3. The path with relatively more pheromone has more chance of being chosen by the ants.

*C. The ACO algorithm for WIL*

Because WIL is also a permutation constrained problem, ACO is directly applicable. In TSP, the ant find a sequence in which to visit cities, while in WIL the ant find an order in which to place items. The main issues concerned in our ACO are listed as follows:

1. The first important step is to define the pheromone trail for the ACO application. In WIL, The pheromone trail $\tau(i, j)$ is defined as the "favorability" of placing item $j$ after item $i$;

2. Another important feature of ACO is the heuristic. In our ACO, the heuristic information $\eta(i, j)$ in (8) is defined as $m_j * r_j$, because heavier and lager items should be placed with higher priority;

3. In our ACO, $L_k$ in (10) is defined as the radius of the enveloping circle of the layout generated by the OPT, so compact layouts are preferred.

4. The update of the pheromone follows the ACO version of Max-Min Ant System (MMAS) (Stützle & Hoos, 1997), where only the ant with the best solution is allowed to deposit pheromone on its trail in each cycle of the iteration. To prevent the ACO converging too quickly, the pheromone is limited in $[\tau_{min}, \tau_{max}]$.

## V. EXPERIMENTS AND RESULTS

We now describe the results of numerical experiments to test our method. The experiments are carried on an Intel 1.83GHZ/512Mb Computer. All algorithms are implemented in the C language and compiled by MS VC++. Each program is run on each benchmark instance 10 times. The minimal radius of the enveloping circle (*r_best*), the average radius of the envelope circle (*r_average*), and the average computation time for each instance (*t_average*) are used to evaluate the performance of the algorithms tested. The computational results are listed in table I and table II.

In what follows, two versions of ACO, the standard Ant System (AS) and the Min-Max AS (MMAS), are implemented and tested. The parameter settings are derived from (Dorigo & Gambardella, 1997). The number of ants is set to 20, and the number of iterations to 100. The importance factors of the pheromone trail $\alpha$ and heuristic $\beta$ are both set to 1. The pheromone decaying rate $\rho$ is set to 0.9. The value of $Q$ in (10) is set to the minimal radius of the enveloping circle found. Before each iteration, pheromone $\tau(i, j)$ is initialized to $1/n$. The $[\tau_{min}, \tau_{max}]$ is set to $[0.1/n, 10/n]$.

### A. Tests on Weighted Circle Layout

We first use 10 problem instances taken from (Xiao *et al.*, 2007a; Xu *et al.*, 2007) to test the ACO for WCL. The number of circles varies from 10 to 55. Two other algorithms are also run on the same benchmark instances, in order to yield meaningful comparisons. One is a local search heuristic from (Xiao *et al.*, 2007a), which is a hybrid of the gradient descent method and Particle Swarm Optimization (PSO), named CA-PSLS (compact algorithm with particle swarm local search). The other one is a GA based on OPT from (Xu *et al.*, 2007). Both algorithms were the best known-methods for this problem at the time of publication.

In CA-PSLS, we first randomly select 100 layouts, and then use the compaction algorithm to get an output layout from each of them. The best layout from the 100 outputs is chosen as the start point, and the PSO algorithm then performs a local search to improve the layout. We use 20 particles and the iteration number of the PSO is set to 5000. In the GA, the population size is set to 20, and the program run for100 generations. One-point crossover is used and the mutation probability set to 12.5%. The computational results for the AS, MMAS, GA, CA-PSLS are listed in Table I.

We derive three findings from these results:

1. MMAS performs better than AS on WCL problems. Because only the best ant updates the pheromone trail in MMAS, while *every* ant in AS updates its own pheromone trail, MMAS runs a little faster than AS on all 10 instances. Only the ant who finds the best solution is permitted to update the pheromone trail, which suggests that MMAS spends more on exploiting than on exploration. So it can be predicted that in larger solution spaces, MMAS will outperform AS. The results also show that on the first two small instances, AS finds better layouts than MMAS, and on the following 8 *larger* instances, MMAS outperforms AS.

2. MMAS search is nearly as effective as the GA on average. In Table I, we see that, on average, on instance 1,2,3,7,8, the GA finds better layouts than MMAS, but on the other five instances, MMAS outperforms the GA. This finding demonstrates that, although ACO and GA are based on different kinds of search idea, if both of them apply heuristic information they can have broadly similar search abilities.

3. OPT is a powerful constructive method. Although the CA-PSLS method yields very successful results on WCL problems compared to other local search methods (Xiao et al., 2007), in Table 1, we find that the three algorithms based on OPT, (AS, MMAS, and GA) yield similar results (in terms of quality) in about 20% of the computational run-time.

TABLE I: Computational Results for the WCL

| instance | algorithm | r_best | r_average | t_average (second) | instance | algorithm | r_best | r_average | t_average (second) |
|---|---|---|---|---|---|---|---|---|---|
| 1 | AS | 60.60966 | 60.94029 | 0.6250 | 6 | AS | 103.7284 | 104.3541 | 20.6375 |
| (10 items) | MMAS | 61.09849 | 61.43261 | 0.5750 | (35 items) | MMAS | 102.2805 | 103.4782 | 19.6390 |
| | GA | 61.32746 | 61.40036 | 0.5844 | | GA | 103.1702 | 104.3123 | 22.1171 |
| | CA-PSLS | 60.98673 | 61.83084 | 4.7390 | | CA-PSLS | 103.5901 | 104.374 | 120.8984 |
| 2 | AS | 67.57098 | 68.83913 | 1.9856 | 7 | AS | 116.3434 | 117.0139 | 30.2578 |
| (15 items) | MMAS | 68.57272 | 69.55288 | 1.8625 | (40 items) | MMAS | 115.902 | 116.9035 | 29.1015 |
| | GA | 67.6733 | 68.87782 | 1.9921 | | GA | 114.254 | 116.7697 | 32.3953 |
| | CA-PSLS | 67.82491 | 69.02639 | 12.1703 | | CA-PSLS | 115.024 | 116.7738 | 160.2546 |
| 3 | AS | 82.95845 | 84.17213 | 4.4281 | 8 | AS | 120.3446 | 121.3257 | 41.8921 |
| (20 items) | MMAS | 82.37723 | 83.89189 | 4.2109 | (45 items) | MMAS | 119.6772 | 120.7255 | 39.5156 |
| | GA | 82.2196 | 83.74096 | 4.5109 | | GA | 119.6579 | 120.5485 | 44.2578 |
| | CA-PSLS | 83.02694 | 84.64698 | 28.5750 | | CA-PSLS | 120.1887 | 120.9147 | 200.9500 |
| 4 | AS | 84.81925 | 85.51662 | 8.1625 | 9 | AS | 126.5541 | 127.548 | 55.6328 |
| (25 items) | MMAS | 83.11417 | 84.41404 | 7.4515 | (50 items) | MMAS | 125.1307 | 126.1234 | 53.3328 |
| | GA | 84.58051 | 85.46078 | 8.4156 | | GA | 125.3068 | 126.5598 | 59.8015 |
| | CA-PSLS | 84.05504 | 85.46442 | 52.2452 | | CA-PSLS | 126.457 | 127.4425 | 235.2781 |
| 5 | AS | 100.5408 | 101.26 | 13.4593 | 10 | AS | 138.6933 | 139.5136 | 73.9078 |
| (30 items) | MMAS | 99.39906 | 100.3329 | 12.2078 | (55 items) | MMAS | 137.4884 | 138.7685 | 70.3265 |
| | GA | 99.80675 | 100.5153 | 13.8765 | | GA | 138.0675 | 138.9203 | 80.0968 |
| | CA-PSLS | 99.96384 | 100.8123 | 86.5687 | | CA-PSLS | 138.9641 | 140.2751 | 279.3734 |

*B. Tests on Weighted Rectangle Layout*

We use the 4 instances in (Xu *et al*., 2007a) to test the ACO for WRL, with 5, 6, 9, 20 rectangles respectively. To test the algorithms on a larger problem, we compose a new instance with 40 rectangles. A local search method, also named CA-PSLS, from (Xu *et al.,* 2007a) is tested as the comparison algorithm. In CA_PSLS, we select one layout randomly and compact it, and use the output as the start point. We run the PSO to perform a local search in order to get a refined layout. The number of particles of PSO is set to 20 and the iteration number to 5000. The results are displayed in Table II. The findings are as follows:

1. We again find that MMAS outperforms AS on the larger instances. On the smaller instances 1, 2, and 3, where the numbers of items are less than 10, AS outperforms MMAS. However, on instances 4 and 5, where the numbers of items grow to 20 and 40 respectively, MMAS wins. This finding shows that MMAS is suitable for searching larger solution spaces.

2. OPT-based MMAS shows clear superiority over the local search method CA-PSLS on large instances. On instances 1 and 2, MMAS finds inferior layouts compared to CA-PSLS, but the differences are very small. When the number of rectangles is raised to 9, CA-PSLS begin to lose ground, and when the number is raised to 40, the radius of the enveloping circle found by MMAS is better than that found by CA-PSLS by a factor of 50% in terms of *r_best*, and by a factor of 100% in terms of *r_average*. Considering computational resources, we find that MMAS was about 3 times faster than CA-PSLS on all instances.

Fig 8 presents the a graphical illustration of the results obtained by both algorithms on instance 5, which can help us to understand the superiority of OPT. We see in Fig 8-b that all the rectangles are adjacent along their edges, according to OPT rules, which seems to be a hard task for CA-PSLS to achieve.

Table II: Computational Results for the WRL.

| instance | algorithm | r_best | r_average | t_average (second) |
|---|---|---|---|---|
| 1 (5 items) | AS | 11.75972 | 11.75972 | 2.859 |
| | MMAS | 11.75972 | 11.75972 | 2.688 |
| | CA-PSLS | 10.9537 | 12.12377 | 12.782 |
| 2 (6 items) | AS | 14.84244 | 14.84244 | 4.562 |
| | MMAS | 15.12812 | 15.32506 | 4.515 |
| | CA-PSLS | 14.39625 | 15.19402 | 19.904 |
| 3 (9 items) | AS | 17.74857 | 18.06303 | 13.406 |
| | MMAS | 18.01857 | 18.38084 | 13.985 |
| | CA-PSLS | 18.60944 | 19.88978 | 53.631 |
| 4 (20 items) | AS | 23.16855 | 23.57499 | 101.500 |
| | MMAS | 22.35156 | 23.15215 | 107.437 |
| | CA-PSLS | 26.40786 | 32.00123 | 311.340 |
| 5 (40 items) | AS | 119.3148 | 122.5032 | 642.562 |
| | MMAS | 117.9112 | 120.2168 | 621.531 |
| | CA-PSLS | 174.6052 | 224.8859 | 1417.300 |

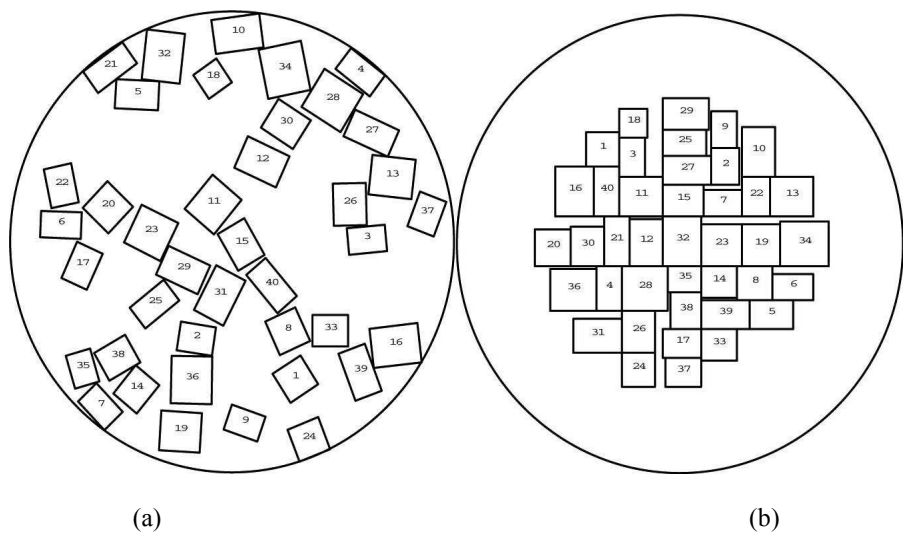

(a)          (b)

Figure 8: The layouts generated from instance 5 in the experiment for WRL. (a) layout by CA-PSLS, (b) layout by MMAS.

## VI. CONCLUSIONS

This paper proposes a constructive heuristic to pack weighted items in a circular container, where the compactness and balance should be considered. By carefully designing packing rules, the heuristics generate good layouts. An ant-based algorithm based on this heuristic is then described to optimize the packing order. In the ant algorithm, the pheromone matrix encodes the favorability of choosing an item. Heuristic information is also considered, which is the product of the size and the weight of the next packed item. This means that large and heavy items have higher priority.

Two versions of the ant-based algorithm, AS and Min-Max AS, are compared with existing approaches, such as the genetic algorithm for Weighted Circle L, and the hybrid particle swarm algorithm for both Weighted Circle and Weighted Rectangle Layout. The experimental results showed that: (1) Min-Max AS performs better than AS on large-scaled instances for both WCL and WRL, (2) The OPT based approaches, including the ACO and GA, perform better than the local search based approach CA-PSLS.

However, the circle and rectangle are two special shapes. Further work is required to study packing of weighted and irregular-shaped items, which will be the subject of future research.

REFERENCES

C. Blum and M. Dorigo (2004). "The hyper-cube framework for ant colony optimization." *IEEE Transactions on System, Man, and Cybernetics, Part B* **34**(2), pp. 1161-1172.

B. Brugger, K.F. Doerner, R.F. Hartl and M. Reimann (2004). "AntPacking – An ant colony optimization approach for the one-dimensional bin packing problem." Lecture Notes in Computer Science, Volume 3004, pp. 41-50.

E. Burke and G. Kendall (1999). "Applying ant algorithms and the no fit polygon to the nesting problem." Lecture Notes in Computer Science, Volume 1747, pp. 453-464.

J. Cagan, D.Degentesh, and S.Yin (1998). "A simulated annealing-based algorithm using hierarchical models for general three-dimensional component layout." *Computer-Aided Design* **30**, pp. 781-790.

M. Dorigo (1992). *Optimization, learning and natural algorithms*, PhD thesis, Politecnico di Milano, Italy.

M. Dorigo, V. Maniezzo and A. Colorni (1996). "Ant system: optimization by a colony of cooperating agents." *IEEE Transactions on System, Man, and Cybernetics, Part B* **26**(1), pp.29-41.

M. Dorigo and L.M. Gambardella (1997). "Ant colony system: a cooperative learning approach to the Travelling Salesman Problem." *IEEE Transactions on Evolutionary Computation* **1**(1), pp.53-66.

M. Dorigo and T. Stützle (2004). *Ant Colony Optimization*, MIT Press.

K A. Dowsland, S. Vaid, and W. B. Dowsland (2002). "An algorithm for polygon placement using bottom-left strategy." *European Journal of Operational Research* **141**(2), pp. 371-381.

H. Dyckhoff (1990). "A typology of cutting and packing problems." *European Journal of Operational Research* **44**(1), pp. 145-159.

J. Gonzalez, I. Rojas, H. Pomares, M. Salmeron and J.J. Merelo (2002). "Web newspaper layout optimization using simulated annealing." *IEEE Transactions on System, Man, and Cybernetics, Part B* **32**(5), pp. 686-691.

J. M. Harwig (2003). *An adaptive tabu search approach to cutting and packing problems.* PhD thesis, University of Texas at Austin.

W. Huang and M. Chen (2006). "Note on: An improved algorithm for the packing of unequal circles within a larger containing circle." *Computers & Industrial Engineering* **50**(3), pp. 338-344.

C.-F. Juang and C.-H. Hsu (2009). "Reinforcement interval type-2 fuzzy controller design by online rule generation and Q-Value-Aided ant colony optimization". *IEEE Transactions on System, Man, and Cybernetics, Part B*, DOI: 10.1109/TSMCB.2009.2020569.

J. Levine and F. Ducatelle (2004). "Ant colony optimization and local search for bin packing and cutting stock problems." *Journal of the Operational Research Society* **55**(7), pp. 705–716.

A. Sokea and Z. Bing (2006). "Hybrid genetic algorithm and simulated annealing for two-dimensional non-guillotine rectangular packing problems." *Engineering Applications of Artificial Intelligence* **19**(5), pp. 557-567.

T. Stützle and H. Hoos (1997). "MAX-MIN ant system and local search for the Traveling Salesman Problem." Proceedings of the 1999 IEEE Congress on Evolutionary Computation, IEEE Press, pp. 309-314.

M. Tang and X. Yao (2007). "A memetic algorithm for VLSI floorplanning." *IEEE Transactions on System, Man, and Cybernetics, Part B* **37**(1), pp. 62-69.

H. Teng, S. Sun, and W. Ge, and W. Zhong (1994). "Layout optimization for the dishes installed on a rotating table- the packing problem with equilibrium behavioral constraints." *Science in China* (*Series A*) **37**(10), pp.1272-1279.

H. Teng, S. Sun, D. Liu, and Y. Li (2001). "Layout optimization for the objects located within a rotating vessel - a three-dimensional packing problem with behavioral constraints." *Computers & Operations Research* **28**(6), pp. 521-535.

D.R. Thiruvady, B. Meyer and A.T. Ernst (2008). "Strip packing with hybrid ACO: placement order is learnable." Proc. IEEE Congress on Evolutionary Computation, IEEE Press, pp.1207-1213.

H. Wang, W. Huang, and Q. Zhang, and D. Xu (2002). "An improved algorithm for the packing of unequal circles within a larger containing circle." *European Journal of Operational Research* **141**(2), pp. 440-453.

Y. L. Wu, W. Huang, S. Lau, C.K. Wong and J.H. Yung (2002). "An effective quasi-human based heuristic for solving the rectangle packing problem." *European Journal of Operational Research* **141**(2), pp. 341-358.

R.-B. Xiao, Y.-C. Xu, and M. Amos (2007). "Two hybrid compaction algorithms for the layout optimization problem." *BioSystems* **90**(2), pp. 560-567.

Y.C. Xu, R. B. Xiao, and M. Amos (2007a). "Particle swarm algorithm for weighted rectangle placement." Proceedings of the 3rd International Conference on Natural Computation (ICNC07), IEEE Press, pp. 728-732.

Y.C. Xu, R. B. Xiao, and M. Amos (2007b). "A novel genetic algorithm for the layout optimization problem." Proceedings of the 2007 IEEE Congress on Evolutionary Computation (CEC07), IEEE Press, pp. 3938-3942.

Y.C. Xu and R. B. Xiao (2008). "An ant colony algorithm for the layout optimization with equilibrium constraints." *Control and Decision* **23**(1), pp. 25-29 (in Chinese).

Y.C. Xu, F. M. Dong, Y. Liu, and R.B. Xiao (2010) "A new genetic algorithm for laying out rectangles with equilibrium constraints." Submitted to *Pattern Recognition and Artificial Intelligence* (In Chinese).

B. Zhang, H.F. Teng and Y.J. Shi (2008). "A layout optimization of satellite module using soft computing techniques." *Applied Soft Computing Journal* **8**(1), pp. 507-521.